\documentclass[letterpaper, 10 pt, conference]{ieeeconf}

\usepackage{amsmath}
\usepackage{amssymb}
\usepackage{mathtools}

\usepackage{tabularx}
\usepackage{graphicx}
\usepackage{bm}
\usepackage{color}
\usepackage{float}
\usepackage{units}
\usepackage{url}
\usepackage{hyperref}
\usepackage{balance}
\usepackage{amsmath}
\usepackage{makecell}
\usepackage{cite}
\usepackage{contour}
\usepackage{xcolor}
\usepackage{amsthm}
\usepackage{algorithm, algorithmicx, algpseudocode}
\usepackage{booktabs, multirow, multicol}
\usepackage[table]{xcolor}
\usepackage[dvipsnames]{xcolor}
\usepackage{adjustbox}
\usepackage{empheq}

\definecolor{lightgrayrow}{RGB}{240,240,240}
\definecolor{lightpinkrow}{RGB}{255,230,235}

\makeatletter 
\def\endfigure{\end@float} 
\def\endtable{\end@float}
\makeatother

\usepackage{subcaption}
\usepackage[]{graphicx}
\usepackage{caption}

\newtheorem{remark}{\textbf{Remark}}
\usepackage{lipsum}
\usepackage{subcaption}

\IEEEoverridecommandlockouts

\begin{document} 

\title{\Large \bf 			
Accelerating and Scaling MPC-Guided Reinforcement Learning for \\Humanoid Locomotion and Manipulation
}


\author{Junheng Li$^{*1}$, Liang Wu$^{*2}$, Sergio A. Esteban$^{1}$, Lizhi Yang$^{1}$, J\'an Drgo\v na$^{2}$, and Aaron D. Ames$^{1}$
\thanks{$^*$Equal contribution; $^{1}$California Institute of Technology, CA 91106, USA; $^{2}$Johns Hopkins University, MD 21218, USA.}\thanks{Corresponding author email:{\tt\small junhengl@caltech.edu}. This work is in part supported by Technology Innovation Institute, The Dow Chemical Company project \#227027AW, Westwood Robotics, the Ralph O’Connor Sustainable Energy Institute at Johns Hopkins University, and by the U.S. DOE, Office of Science, ASCR program under the Scientific Discovery through Advanced Computing (SciDAC) Institute “LEADS: LEarning-Accelerated Domain Science”.}
}	
\maketitle


\begin{abstract}
In humanoid motion control, model predictive control (MPC) offers physically grounded prediction and constraint handling, while reinforcement learning (RL) enables robust whole-body skills through large-scale simulation. However, using MPC inside RL often requires time-consuming problem construction or excessive training overhead, making such frameworks difficult to justify in practice. This work studies efficient training-time MPC guidance for humanoid locomotion and manipulation, termed \textit{MPC-RL}. We introduce a centroidal-dynamics MPC reward formulation that leverages guidance from MPC trajectories in training time. To make this practical in massively parallel RL, we develop $\pi^n$MPC, a \underline{p}arallel-\underline{i}n-horizon and construction-free batched GPU MPC solver that operates directly on time-varying dynamics to avoid high memory usage and pre-compilation. Through a variety of comparative studies and hardware validations, we have found that MPC-RL achieves superior performance in locomotion and manipulation skills. 
The code base is available at \url{https://github.com/junhengl/mpc-rl}

\end{abstract}


\section{Introduction}
\label{sec:Introduction}

Recent advances in humanoid locomotion and manipulation control and learning have opened broad opportunities for operation in human-centered environments \cite{gu2026humanoid}. Model predictive control (MPC) and reinforcement learning (RL) represent two complementary paradigms for humanoid motion control. MPC provides explicit dynamics modeling, constraint handling, and predictive reasoning \cite{khazoom2024tailoring, romualdi2022online, wensing2023optimization}, while RL has recently demonstrated impressive robustness and agility \cite{li2021reinforcement, dao2024sim, liu2025opt2skill}. Therefore, combining MPC and RL has emerged as a promising direction for humanoid control \cite{kamohara2025rl, bang2024rl, jeon2025residual}.

Recent work has explored RL-augmented MPC for legged and humanoid control, \cite{chen2024learning, kamohara2025rl, jeon2025residual} train residual policies and greatly enhance the capability of MPC in deployment. However, such frameworks require efficient test-time MPC solutions and still rely on real-time state estimation. Parallelly, offline optimized trajectory libraries also provide dynamically consistent guidance for RL \cite{liu2025opt2skill, li2025amo} and avoid excessive computation during training. These frameworks allow dynamic and interactive motion control; however, the exploration is limited to the motion encoded offline. Additionally, there is an increasing interest in the direction of physics-based structure and guidance in RL. For example, Dai \textit{et al.} \cite{dai2026walk} leverages analytical gait synthesis with CLF-based reward structure for future foot placement guidance in highly unstructured terrain. Krishna \textit{et al.} \cite{krishna2024ogmp} uses interpolated oracle guidance for multi-modal locomotion.
In short, predictive guidance can provide dynamically consistent motion targets, contact-aware structure, and long-horizon task information that are often difficult to recover from reward design alone. 

In the aerial robotics community, training-time MPC-guided learning has shown strong promise, where MPC can provide physically structured supervision for learned policies \cite{romero2024actor, greatwood2019reinforcement}. Zhang \emph{et al.} \cite{zhang2016learning} used MPC-guided policy search to train quadrotor policies from full-state MPC rollouts, enabling lower-cost onboard execution at test time. 

\begin{figure}[!t]
    \center	
    \includegraphics[clip, trim=0.0cm 0.0cm 0.0cm 0.0cm, width=1\columnwidth]{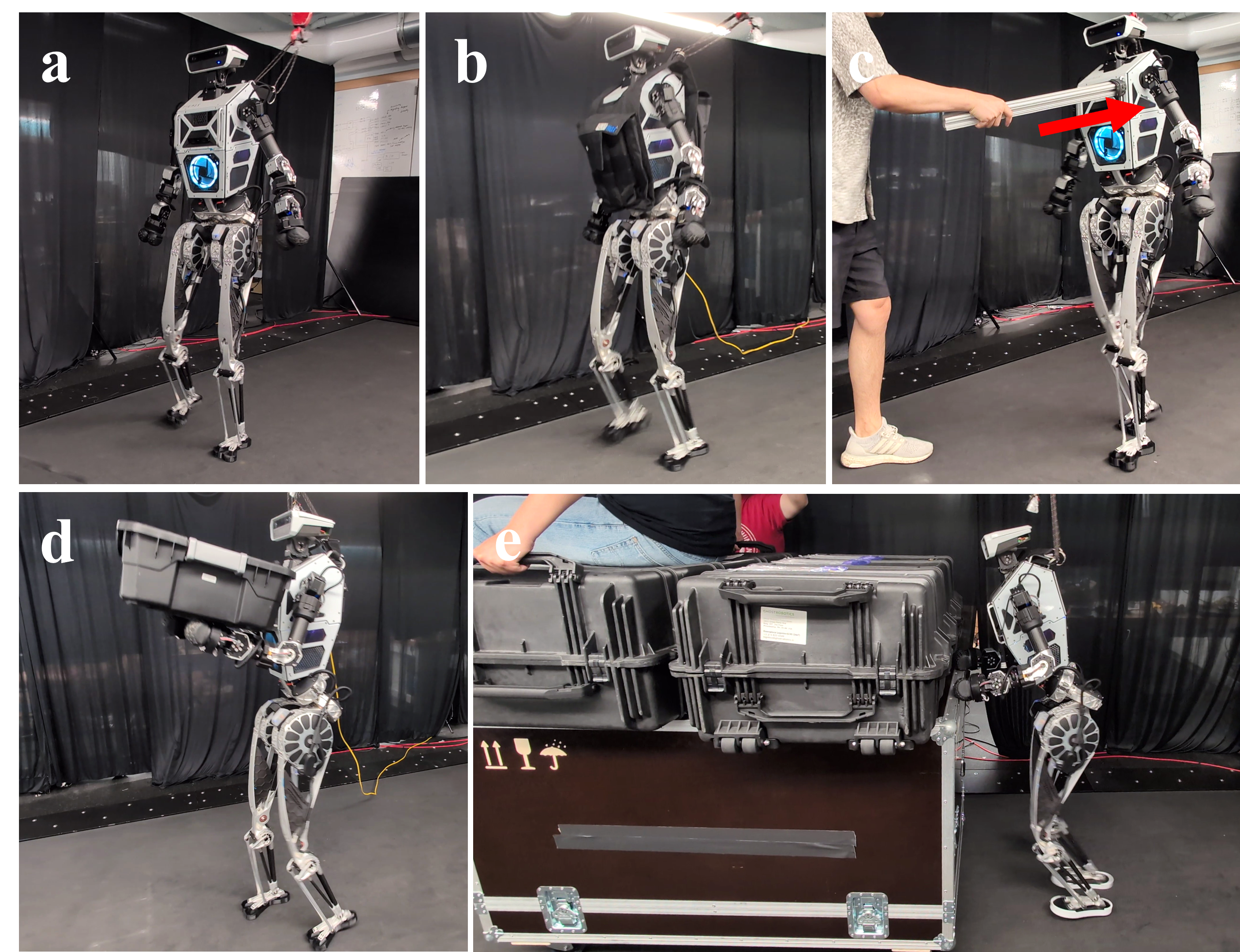}
    \caption{\textbf{MPC-RL Experiment Snapshots.} a). Walking on treadmill; b). Wearing an unknown 10 kg vest while walking; c). Push-recovery; d). Carrying an unknown 10 kg payload while walking; e). Pushing a 290kg cart. Full video: \url{https://youtu.be/PrcbXkA1kYg} }
    \label{fig:hardware}
    \vspace{-0.4cm}
\end{figure}

\begin{figure*}[!t]
\vspace{0.2cm}
    \center	
    \includegraphics[clip, trim=0.0cm 0.0cm 0.0cm 0.0cm, width=2\columnwidth]{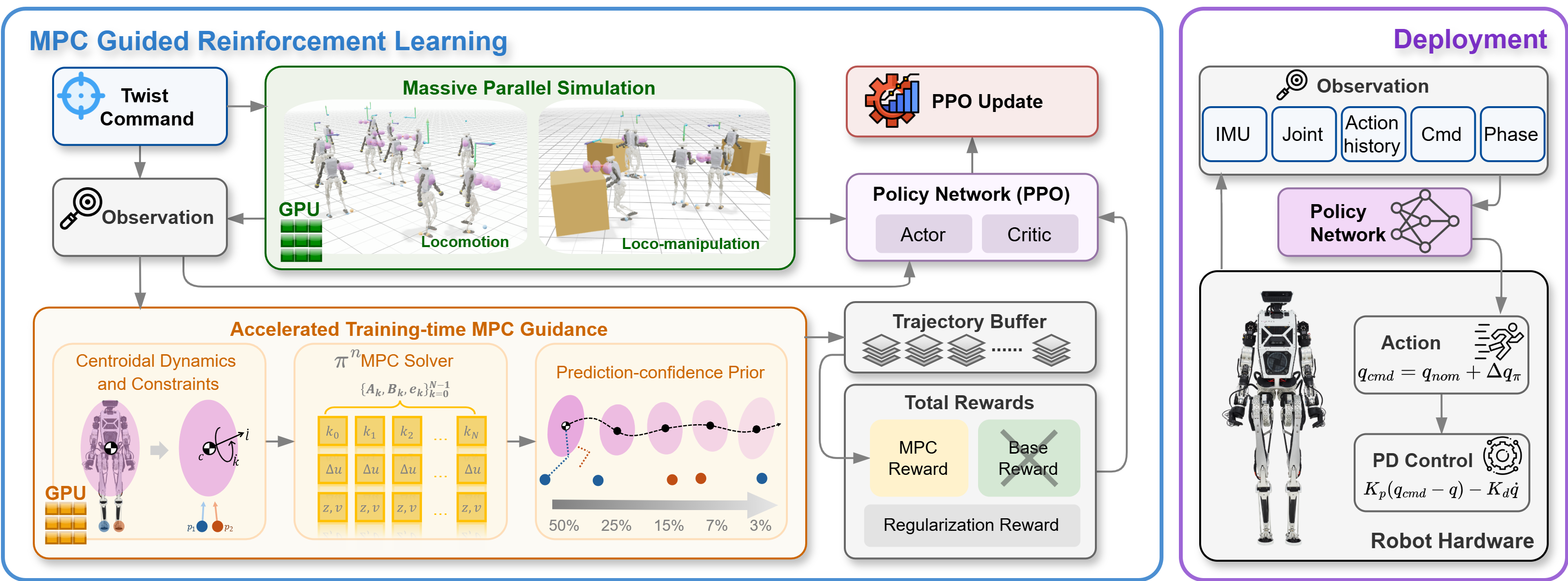}
    \caption{MPC-Guided Reinforcement Learning (MPC-RL) Architecture. During training, MPC generates per-environment predictive references through $\pi^n$MPC batched solver, which are converted into prediction landmark rewards for PPO; at deployment, the learned policy deploys without MPC.}
    \label{fig:system}
    \vspace{-0.4cm}
\end{figure*}

Extending long-horizon MPC guidance to humanoid RL remains challenging: repeated batched MPC solves force a tradeoff between prediction length, update rate, and training time \cite{jenelten2024dtc,reiter2026synthesis}. 
Recent GPU-based MPC solvers reduce the cost of running many MPC instances in parallel during RL training \cite{jeon2024cusadi, bishop2024relu}. However, existing approaches often parallelize across environments more naturally than across the prediction horizon, leaving horizon-wise recursions or sequential optimization steps as a bottleneck. Traditionally, forming MPC as sparse QP problems scales poorly with problem size in VRAM usage and in symbolic compilation resources, requiring time-consuming problem construction \cite{kamohara2025rl}. Therefore, a \textit{construction-free} and \textit{memory-efficient} batched solver is highly needed.
Additionally, most prior studies incorporating this MPC$\leftrightarrow$RL hybrid approach focus primarily on locomotion \cite{cai2026cost}, while MPC-guided RL for loco-manipulation remains under-explored, where physical and interactive models are higly valuable \cite{liu2025opt2skill}.

In this work, we study the role of training-time MPC guidance in RL for humanoid locomotion and loco-manipulation. The core architecture is shown in Figure \ref{fig:system}.
In this work, we make three main contributions. 
(1) We present a structured MPC-based reward for RL, using prediction-confidence weighting to extract long-horizon motion and dynamics guidance from centroidal MPC trajectories, and we provide a systematic empirical study of its effect through ablations on horizon length, update rate, and reward structure.
(2) We introduce $\pi^n$MPC, a \underline{p}arallel-\underline{i}n-horizon, construction-free batched MPC solver that extends the original monolithic and time-invariant $\pi$MPC structure \cite{wu2026pi} to legged robots with time-varying dynamics matrices, enabling compilation-free and memory-efficient batched solves over long horizons. 
(3) We extend the MPC-guided RL architecture beyond locomotion to humanoid loco-manipulation, demonstrating box-pushing on hardware with up to 290 kg (639 lbs) cart payload, corresponding to 829$\%$ of the robot body mass.
Together, these results show how powerful predictive optimization can serve both as a practical training tool and as a source of structure for learning humanoid behaviors with competitive training efficiency.

\section{Centroidal-Dynamics MPC Guidance}
\label{sec:cd_mpc_guidance}

This section describes the centroidal-dynamics \cite{orin2013centroidal} MPC (CD-MPC) module used to provide predictive guidance during RL. The purpose of CD-MPC is not to directly control the humanoid at deployment, but to generate CoM, momentum, ground reaction force (GRF), and footstep references during massive-parallel training as structured rewards. 

\subsection{Centroidal Dynamics Model}
Let $q \in \mathbb{R}^{n_q}$ and $\dot q \in \mathbb{R}^{n_q}$ denote the full-order humanoid configuration and velocity. We define the centroidal state as
\begin{equation}
    \xi =
    \begin{bmatrix}
        c^\top & l_G^\top & k_G^\top
    \end{bmatrix}^\top \in \mathbb{R}^{9},
\end{equation}
where $c \in \mathbb{R}^3$ is the center-of-mass (CoM) position, $l_G \in \mathbb{R}^3$ is the centroidal linear momentum, and $k_G \in \mathbb{R}^3$ is the centroidal angular momentum about the CoM.

Based on the continuous-time CD \cite{orin2013centroidal}, using timestep $\Delta t$, the discrete CD are
\begin{align}
    c_{k+1} &= c_k + \Delta t \frac{1}{m} l_{G,k}, \label{eq:cd_discrete_com}
\end{align}
\begin{align}
    l_{G,k+1} &= l_{G,k} + \Delta t
    \big(
        mg + \sum_{i \in \mathcal{C}_k} f_{i,k}
    \big), \label{eq:cd_discrete_linear}
\end{align}
\begin{align}
    k_{G,k+1} &= k_{G,k} + \Delta t
    \sum_{i \in \mathcal{C}_k}
    \left(
        (p_{i,k} - c_k) \times f_{i,k}
        + \tau_{i,k}
    \right), \label{eq:cd_discrete_angular}
\end{align}
where $m$ is the total robot mass, $g$ is the gravity vector, $\mathcal{C}_k$ is the active contact set at step $k$, $p_{i,k}$ is the position of contact $i$, and $(f_{i,k},\tau_{i,k})$ are the corresponding contact force and moment. For locomotion, $\mathcal{C}_k$ contains the foot contacts planned by Raibert heuristics; for loco-manipulation, it can additionally include hand contacts to reason external interaction forces with environment.
\subsection{CD-MPC Formulation}
At planning time $t$, CD-MPC solves a finite-horizon optimal control problem over $N$ steps, with horizon length
\begin{equation}
    H = N \Delta t.
\end{equation}
The decision variables are the centroidal states and contact wrenches
\begin{equation}
    \mathcal{Z}
    =
    \left\{
        \xi_k, f_{i,k}, \tau_{i,k}
    \right\}_{k=0}^{N-1}.
\end{equation}
The initial centroidal state is obtained from the current full-order simulation state, $\xi_0 = \phi(q_t,\dot q_t)$,
where $\phi$ maps the full-order state to the centroidal state.

The CD-MPC problem is formulated as
\begin{subequations}\label{eq:cd_mpc_problem}
\allowdisplaybreaks
    \begin{align}
    \min_{\mathcal{Z}} \quad
    &
    \sum_{k=0}^{N-1}
    L_k(\xi_k, u_k)
    +
    L_N(\xi_N)
    \\
    \mathrm{s.t.} \quad
    &
    \xi_{k+1}
    =
    f_{\Delta t}(\xi_k,u_k),
    \qquad k = 0,\dots,N-1,
    \\
    &
    f_{i,k} \in \mathcal{F}_{i,k},
    \qquad
    \tau_{i,k} \in \mathcal{T}_{i,k},
    \qquad
    i \in \mathcal{C}_k,
    \\
    &
    \xi_0 = \phi(q_t,\dot q_t),
    \end{align}
\end{subequations}

where $u_k := \{f_{i,k},\tau_{i,k}\}_{i\in\mathcal{C}_k}$ and $f_{\Delta t}$ denotes the discrete CD from \eqref{eq:cd_discrete_com}--\eqref{eq:cd_discrete_angular}. $\mathcal{F}_{i,k}$ and $\mathcal{T}_{i,k}$ denote the constraint sets based on contact wrench cone (CWC) \cite{caron2015stability} and friction pyramid.

The stage and terminal cost contains velocity tracking, CoM tracking, angular momentum regulation, and control input regularization:
\begin{align}
    L_k
    =
    &
    \|v^c_k - v^{\mathrm{cmd}}_t\|^2_{Q_v}
    +
    \|c_k - c^{\mathrm{ref}}_k\|^2_{Q_c}
    +
    \|k_{G,k} - k_{G,k}^{\mathrm{ref}}\|^2_{Q_{k_G}}
    \notag \\
    &
    +
    \sum_{i\in\mathcal{C}_k}
    \|u_{i,k}\|^2_{R_u},
    \label{eq:cd_mpc_stage_cost}
\end{align}
where $v^c_k = \frac{1}{m} l_{G,k}$.
For nominal locomotion, $k_{G,k}^{\mathrm{ref}}$ and $c^{\mathrm{ref}}_k$ are chosen based on the reference structure proposed in \cite{li2025gait}.

\subsection{MPC Landmark Guidance Reward}
\label{sec:mpc_landmark_reward}

Solving~\eqref{eq:cd_mpc_problem}, it returns a predicted centroidal trajectory
\begin{equation}
    \hat{\xi}^{\,t}(\delta)
    =
    \begin{bmatrix}
        \hat c^{\,t}(\delta)^\top &
        \hat l_G^{\,t}(\delta)^\top &
        \hat k_G^{\,t}(\delta)^\top
    \end{bmatrix}^\top,
    \:
    \delta\in[0,H].
\end{equation}
A naive use of this trajectory is to track only the next MPC prediction.
However, this discards the longer-horizon structure of MPC. We instead extract a small set of predictive landmarks from the full trajectory. Let
\begin{equation}
    \delta_\ell = n_\ell\Delta t, \qquad
    n_\ell = 1+\lfloor s_\ell(N-1)\rfloor,
    \qquad
    s_\ell\in[0,1],
\end{equation}
where $\ell=0,\dots,N_L-1$, $s_0=0$, and $s_{N_L-1}=1$.
Thus, $\delta_0$ corresponds to the next MPC step and
$\delta_{N_L-1}=H$ corresponds to the end of the horizon.

For each MPC solve at time $t$, we store the landmarks
\begin{equation}
    \left\{
    \left(t+\delta_\ell,\hat c^{\,t}(\delta_\ell)\right)
    \right\}_{\ell=0}^{N_L-1}.
\end{equation}
When the rollout reaches time $t$, the realized CoM can be compared
against the landmarks predicted for this same time from previous MPC
solves. We define
\begin{equation}
    e_c^\ell(t)
    =
    c_t - \hat c^{\,t-\delta_\ell}(\delta_\ell),
\end{equation}
where unavailable landmarks at the beginning of an episode are omitted.
The CoM MPC-guidance reward is then
\begin{equation}
    r_{t}^{\mathrm{mpc},c}
    =
    \exp\!\left(
    -\sum_{\ell=0}^{N_L-1}
    \alpha_\ell
    \frac{\|e_c^\ell(t)\|_{Q_c}^{2}}{\sigma_c^2}
    \right).
    \label{eq:mpc_com_landmark_annealing}
\end{equation}
The weights follow a horizon-monotone schedule
\begin{equation}
    \alpha_0 > \alpha_1 > \cdots > \alpha_{N_L-1} > 0,
\end{equation}
which we refer to as \emph{prediction-confidence weighting}. The schedule
encodes a prediction-confidence prior: near-horizon landmarks are weighted
more heavily because they are less affected by centroidal model mismatch and integration error, while far-horizon landmarks are down-weighted.

\begin{remark}
The single exponential form in~\eqref{eq:mpc_com_landmark_annealing}
fuses all landmark residuals into one bounded reward
$r_t^{\mathrm{mpc},c}\in(0,1]$. 
Two limiting cases recover familiar designs: if $\alpha_0=1$ and $\alpha_{\ell>0}=0$, the reward reduces to strict next-step MPC supervision; if $\alpha_\ell=1/N_L$, it becomes a time-averaged trajectory-tracking objective. In all of our MPC-RL experiments, we set the confidence-weighted schedule to a horizon-tapered profile $\alpha = \{0.5,~0.25,~0.15,~0.07,~0.03\}$, allowing the policy to exploit long-horizon MPC structure while keeping the near-term tracking signal sharp.
\end{remark}

The same landmark construction is used for other MPC-guided quantities, including CoM velocity and angular momentum references. The total PPO reward is
\begin{equation}
    r_t
    =
    r_t^{\mathrm{reg}}
    +
    \eta\, r_t^{\mathrm{mpc}},
\end{equation}
where $r_t^{\mathrm{reg}}$ contains the standard regularization rewards and $r_t^{\mathrm{mpc}}$ collects the CD-MPC landmark guidance terms, which is covered in more detail in Section \ref{sec:training-setup}.

\section{$\pi^n$MPC: \underline{P}aralle-\underline{i}n-horizon and Construction-free Batched MPC Solver}
\label{sec:piTorch_mpc}

To enable MPC solving effort and guidance lightweight during RL training, we build on top of the parallel-in-horizon ADMM structure of previous work $\pi$MPC~\cite{wu2026pi}. Unlike the native Julia implementation $\pi$MPC with time-invariant state matrices for monolithic MPC solve, we adapt to the time-varying nature of the legged locomotion dynamics and allow storage of time-varying state matrices along the horizon for massive-parallel evaluations. The implementation is migrated and realized in both PyTorch and JAX, termed \emph{$\pi^n$MPC}. At every MPC step, we solve a  linearized CD-MPC in the standard form following eqn. (\ref{eq:cd_mpc_problem}),
\begin{align}
\min_{\xi_{1:N},u_{0:N-1}} \quad
&\sum_{k=0}^{N-1}
\frac{1}{2}\|C\xi_{k+1}-r^y_k\|_{Q_k}^2
+\frac{1}{2}\|u_k-r^u_k\|_{R_k}^2  \nonumber\\
\mathrm{s.t.}\quad
&\xi_{k+1}=A_k\xi_k+B_ku_k+e_k,\:
\xi_0=\phi(q_t,\dot q_t),                            \label{eq:cd_ltv_mpc}\\
&\xi_{k+1}\in\mathcal X_k,\qquad u_k\in\mathcal U_k ,
\end{align}
where $\xi_k$ is the centroidal state defined in Sec.~\ref{sec:cd_mpc_guidance} and $u_k$ denotes contact wrenches. The state matrices $\{A_k,B_k,e_k\}_{k=0}^{N-1}$ are obtained along the input periodic contact sequence.

Unlike previous ADMM-based MPC approaches, $\pi^n$MPC achieves parallel-in-horizon and construction-free while maintaining simple, low-cost operation at each iteration by employing a velocity-form MPC and a novel variable-splitting scheme. \textbf{(1) The velocity-form MPC scheme:} we optimize over input increments.
Define the lifted state
\begin{equation}
    \bar \xi_k =
    \begin{bmatrix}
        \xi_k \\ u_{k-1}
    \end{bmatrix},\qquad
    \bar \xi_{k+1}
    =
    \bar A_k \bar \xi_k
    +
    \bar B_k \Delta u_k
    +
    \bar e_k ,
\end{equation}
with
\begin{equation}
    \bar A_k =
    \begin{bmatrix}
        A_k & B_k\\
        0 & I
    \end{bmatrix},
    \quad
    \bar B_k =
    \begin{bmatrix}
        B_k\\
        I
    \end{bmatrix},
    \quad
    \bar e_k =
    \begin{bmatrix}
        e_k\\
        0
    \end{bmatrix}.
\end{equation}
The lifted constraints are written compactly as
$\bar \xi_{k+1}\in\bar{\mathcal X}_k$, where
$\bar{\mathcal X}_k=\mathcal X_k\times\mathcal U_k$. \textbf{(2) The proposed variable-splitting scheme:} we introduce auxiliary variables
$z_{k+1}= \bar \xi_{k+1}$ and
$v_k= \bar B_k\Delta u_k$, yielding the following optimization problem,
\begin{subequations}\label{eqn_pi_MPC}
\allowdisplaybreaks
\begin{align}
\min~
&\sum_{k=0}^{N-1}
\frac{1}{2}\bar\xi_{k+1}^\top \bar Q_k \bar\xi_{k+1}
-\bar q_k^\top \bar\xi_{k+1} + \Pi_{\bar{\mathcal{X}}_{k+1}}(z_{k+1}) \\
\mathrm{s.t.}~&\bar\xi_{k+1}-z_{k+1}=0,~\forall k\in\mathbb{N}_{0}^{N-1},\label{eqn_pi_MPC_b}\\
&\bar B_k\Delta u_k-v_k=0,~\forall k\in\mathbb{N}_{0}^{N-1},\label{eqn_pi_MPC_c}\\
&z_{k+1}-\bar A_k\bar\xi_k-v_k-\bar e_k=0,~\forall k\in\mathbb{N}_{0}^{N-1}.\label{eqn_pi_MPC_d}
\end{align}   
\end{subequations}
where the indicator function $\Pi_{\bar{\mathcal{X}}_{k+1}}(z_{k+1})$ denotes the constraint set $z_{k+1}\in\bar{\mathcal{X}}_{k+1}$. \textit{Note that the above splitting scheme is the key to achieving linear parallel-in-horizon and construction-free features, such as making a copy of $v_k=\bar B_k\Delta u_k$ instead of $v_k=\Delta u_k$, and incorporating $z_{k+1}$, rather than $\bar \xi_{k+1}$, into the dynamical equation.}

The augmented Lagrangian (with the scaled-form Lagrangian variable) of Problem \eqref{eqn_pi_MPC} is given by
\[
\begin{aligned}
\mathcal{L}_{\rho}&=\sum_{k=0}^{N-1}
\frac{1}{2}\bar\xi_{k+1}^\top \bar Q_k \bar\xi_{k+1}
-\bar q_k^\top \bar\xi_{k+1} + \Pi_{\bar{\mathcal{X}}_{k+1}}(z_{k+1}) \\
&+ \frac{\rho}{2}\sum_{k=0}^{N-1}\|\bar\xi_{k+1}-z_{k+1}+\theta_k \|_2^2 + \| B_k\Delta u_k-v_k + \beta_k\|_2^2\\    
&+ \frac{\rho}{2}\sum_{k=0}^{N-1} \|z_{k+1}-\bar A_k\bar\xi_k-v_k-\bar e_k + \lambda_k \|_2^2
\end{aligned}
\]
where $(\theta_k,\beta_k,\lambda_k)$ denotes the scaled dual variables for equality constraints \eqref{eqn_pi_MPC_b}, \eqref{eqn_pi_MPC_c}, and \eqref{eqn_pi_MPC_d}, respectively; $\rho$ denotes the ADMM parameter. In our variable-splitting scheme, the three block variable ADMM updates are 
\begin{subequations}\label{eqn_ADMM_update}
\begin{empheq}[left=\empheqlbrace]{align}
    &\textit{update }\big(\{\Delta u\}_{0:N-1}, \{\bar \xi\}_{1:N}\big)\\
    &\textit{update }\big(\{z\}_{1:N}, \{v\}_{0:N-1}\big)\\
    &\textit{update }\big(\{\theta\}_{0:N-1}, \{\beta\}_{0:N-1}, \{\lambda\}_{0:N-1}\big).
\end{empheq}    
\end{subequations}
Specifically, by precomputing/caching $J_k=(\bar B_k^\top\bar B_k)^{-1}\bar B_k^\top$ (note that $\Bar{B}_{t,k}^\top\Bar{B}_{t,k}=B_{t,k}^\top B_{t,k}+I\succ0$),
\[
H_k\triangleq\left\{ \begin{array}{l}
                (\Bar{Q}+\rho I)^{-1},~\text{if }k=N-1\\
                (\Bar{Q}+\rho I+\rho\Bar{A}_{t,k+1}^\top \Bar{A}_{t,k+1})^{-1},~\text{else}           
            \end{array}\right. 
\]
before ADMM iterations, the above updates \eqref{eqn_ADMM_update} result in the following closed-form horizon-parallel updates at $(m+1)$-th ADMM iteration:
\begin{subequations}
\begin{empheq}[left=\empheqlbrace]{align}
&\textbf{for } k=0,\cdots, N-1~\textbf{(in parallel)}\nonumber\\
    &\quad \Delta u_k^{m+1}=J_k(v_k^m-\beta_k^m),\nonumber\\
    &\quad \bar\xi_{k+1}^{m+1}=H_k \tilde h_k,\nonumber\\
&\textbf{for } k=0,\cdots, N-1~\textbf{(in parallel)}\nonumber\\
&\quad \gamma_k=\bar B_k\Delta u_k^{m+1}+\beta_k^m-\bar A_k\bar\xi_k^{m+1} -\bar e_k +\lambda_k^m,\nonumber \\
&\quad \tilde \gamma_k=\bar\xi_{k+1}^{m+1}+\theta_k^m+\bar A_k\bar\xi_k^{m+1}+\bar e_k-\lambda_k^m\nonumber\\
&\quad z_{k+1}^{m+1}=\mathrm{Proj}_{\bar{\mathcal X}_k}
\left(\frac{2\tilde \gamma_k+\gamma_k}{3}
\right),\nonumber\\
&\quad v_k^{m+1}=\frac{1}{2}(z_{k+1}^{m+1}+\gamma_k),\nonumber\\
&\textbf{for } k=0,\cdots, N-1~\textbf{(in parallel)}\nonumber\\
&\quad \theta_k^{m+1}=\theta_k^m+\bar\xi_{k+1}^{m+1}-z_{k+1}^{m+1}, \nonumber\\
&\quad \beta_k^{m+1}=\beta_k^m+\bar B_k\Delta u_k^{m+1}-v_k^{m+1}, \nonumber\\
&\quad \lambda_k^{m+1}=\lambda_k^m+z_{k+1}^{m+1}
-\bar A_k\bar\xi_k^{m+1}-v_k^{m+1}-\bar e_k .\nonumber    
\end{empheq}
\end{subequations}
where 
{
\footnotesize
\[
\tilde h_k\triangleq\left\{\begin{array}{l}
            \bar{q}+\rho(z_{k+1}^m-\theta_k^m),~ \text{if }k=N-1\\
                 \bar{q}+\rho\Big(z_{k+1}^m-\theta_k^m+\Bar{A}_{t,k+1}^\top(z_{k+2}^m-v_{k+1}^m-\Bar{e}_{k}+\lambda_{k+1}^m)\Big),\text{else}
            \end{array} \right.
\]
}

The proof and the detailed derivations (such as the update of $z_{k+1}$) can be found in our previous work \cite{wu2026pi}.

\begin{remark}
$\pi^n$MPC exposes the prediction horizon as a batch dimension, so the local ADMM updates at all horizon indices can be executed in parallel on the GPU rather than through a sequential Riccati-style recursion. In addition, the
solver is construction-free: it operates directly on the time-varying dynamics $\{A_k,B_k,e_k\}_{k=0}^{N-1}$ and avoids assembling a sparse MPC-QP at every update. This is particularly useful for RL training to avoid high VRAM usage or problem-specific compilation or symbolic code generation ahead of time (\textit{e.g.}, \cite{jeon2024cusadi}), $\pi^n$MPC is implemented with standard batched operations and can adapt online.
\end{remark}
\begin{remark}
    Furthermore, to accelerate the convergence speed, the $\pi^n$MPC implementation inherits Nesterov's acceleration scheme with the restart option, whose details can be found in previous work \cite{goldstein2014fast, wu2026pi}.
\end{remark}


\section{MPC-guided Reinforcement Learning}
\label{sec:training-setup}

This section presents the implementation details of the MPC-RL training and deployment infrastructure. 

\subsection{Training Setup}
We train the policy with PPO using
\texttt{rsl-rl} on top of \texttt{mjlab} \cite{zakka2026mjlab}, running $N_{\mathrm{env}}\!=\!4096$ environments.  Physics is integrated
at $200\,$Hz with with a control rate of $50\,$Hz.  Episodes last at most $T_{\max}\!=\!20\,$s (1000 control steps). The action $a_t\!\in\!\mathbb{R}^{29}$ is a joint-position
delta around the default pose $q^\star$, with action scale $0.5\,$ and tracked by
per-joint PD controllers; the actor's policy is therefore directly
deployable on hardware \textit{without} the MPC in the loop at the test time.  The actor and critic are independent MLPs of width $(512,\,256,\,128)$ with ELU activations and a Gaussian action distribution with a learnable scalar standard deviation.  

\subsection{Policy Observations}
We adopt an \emph{asymmetric actor--critic} formulation: the actor observes only proprioception and the joystick command, so the trained policy is hardware-deployable: trunk angular velocity from the IMU ($\mathbb{R}^{3}$), projected gravity ($\mathbb{R}^{3}$), joint positions and velocities relative to the default pose ($\mathbb{R}^{29}\!\!\times\!\mathbb{R}^{29}$), the previous action ($\mathbb{R}^{29}$), the velocity command $(v_x^{c},v_y^{c},\omega_z^{c})$, and a two-dimensional phase clock $(\sin\phi,\cos\phi)$ that locks the policy to the nominal gait schedule.  All actor inputs are corrupted by uniform observation noise (e.g.\ $\pm 0.2\,$rad/s on body angular velocity, $\pm 0.05$ on projected gravity, $\pm 0.01\,$rad on joint positions, $\pm 1.5\,$rad/s on joint velocities) to encourage robustness to sensor noise during deployment.  The critic is privileged with the same proprioceptive vector \emph{without} noise, plus base linear velocity, per-foot clearance height, and -- specific to MPC-RL -- the MPC's predicted CoM and angular-momentum references $c^{\mathrm{mpc}}_{t},\,k_{G,t}^{\mathrm{mpc}}\!\in\!\mathbb{R}^{3}$, so the value head sees the same MPC prior that shapes the rewards.

\subsection{Reward Structure}

\definecolor{BaseRL}{RGB}{235,248,235}
\definecolor{BaseRLHead}{RGB}{200,235,200}
\definecolor{MPCReward}{RGB}{255,244,226}
\definecolor{MPCRewardHead}{RGB}{255,222,176}
\definecolor{RegReward}{RGB}{230,230,230}
\definecolor{RegRewardHead}{RGB}{200,200,200}

\begin{table}[h]
\vspace{0.15cm}
\centering\footnotesize
\caption{Locomotion Training Reward Structure.}
\label{tab:reward_combined}
\setlength{\tabcolsep}{2pt}
\renewcommand{\arraystretch}{1.18}
\begin{tabularx}{\columnwidth}{@{}l >{\raggedright\arraybackslash}X c@{}}
\toprule
Term & Expression & Weight \\
\midrule

\rowcolor{BaseRLHead}
\multicolumn{3}{@{}l@{}}{\textbf{Base-RL rewards} (Only appear in MPC-free baseline training) } \\
\texttt{lin\_vel}
& $\exp(-\|e_v\|^2/0.25)$
& $+2.0$ \\
\texttt{ang\_vel}
& $\exp(-(e_\omega^2+\|\omega_{xy}\|^2)/0.5)$
& $+2.0$ \\
\texttt{ang\_vel\_reg}
& $-\|\omega_{xy}\|^2$
& $-5{\times}10^{-2}$ \\
\texttt{ang\_mom\_reg}
& $-\|k\|^2$
& $-2{\times}10^{-2}$ \\
\texttt{foot\_slip}
& $-\sum_i\|\dot p_i^{xy}\|^2\,1[\mathrm{ctc}_i]\,g$
& $-0.10$ \\

\midrule
\rowcolor{MPCRewardHead}
\multicolumn{3}{@{}l@{}}{\textbf{MPC rewards} (Only appear in MPC-RL training)} \\
\texttt{mpc\_com}
& $\exp(-\sum_j \alpha_j\|e_c^j\|^2),\: j = 1,2,\dots 5$
& $+2.0$ \\
\texttt{mpc\_lin\_vel}
& $\exp(-\sum_j \alpha_j\|e_{\dot c}^{\,j}\|^2/0.25),\: j = 1,2,\dots 5$
& $+2.0$ \\
\texttt{mpc\_ang\_mom}
& $-\sum_j\alpha_j(\|e_k^j\|^2/0.5+0.01\|\dot e_k^j\|^2)$
& $+5{\times}10^{-2}$ \\
\texttt{mpc\_foot}
& $\mathrm{mean}_{i\in\mathcal S}\exp(-\|p_i^{xy}-p_i^m\|^2/0.04)$
& $+1.0$ \\
\texttt{mpc\_grf}
& $-\sqrt{\frac{1}{6}\sum_j(f_j-f_j^m)^2}$
& $+2{\times}10^{-3}$ \\

\midrule
\rowcolor{RegRewardHead}
\multicolumn{3}{@{}l@{}}{\textbf{Regularization rewards} (Appear in all trainings)} \\
\texttt{pose}
& $\exp(-\overline{(q-q^*)^2/\sigma^2(v^c)}-\|g_{xy}^b\|^2/0.2)$
& $+1.0$ \\
\texttt{gait}
& $\mathrm{mean}_i\,1[\mathrm{ctc}_i=s_i(\phi)]$
& $+1.0$ \\
\texttt{foot\_clear}
& $-\sum_i |h_i-h^*|\,\|\dot p_i^{xy}\|\,g$
& $-4.0$ \\
\texttt{stand\_still}
& $-\|q-q^*\|^2\,1[\|v^c\|<0.1]$
& $-4.0$ \\
\texttt{action\_rate}
& $-\|a_t-a_{t-1}\|^2$
& $-0.25$ \\
\texttt{foot\_height}
& $-\sum_i (h_i^{\mathrm{pk}}/h^*-1)^2\,1_i$
& $-0.25$ \\
\texttt{joint\_lim}
& $-\sum_j([q_j-q_j^{\max}]_+ + [q_j^{\min}-q_j]_+)$
& $-1.0$ \\
\texttt{self\_coll.}
& $-\sum_{c\in g}1[\|f_c\|>10\,\mathrm{N}]$
& $-1.0^\dagger$ \\
\texttt{soft\_land}
& $-\sum_i\|f_i\|\,1[\mathrm{land}_i]$
& $-10^{-5}$ \\

\bottomrule
\multicolumn{3}{@{}l@{}}{\scriptsize
$^\dagger$ feet--feet, hip--abad, hip--hand, foot--leg.}
\end{tabularx}
\vspace{-0.3cm}
\end{table}

The complete reward structure is listed in
Table~\ref{tab:reward_combined}.  The locomotion-driving signal is the MPC stack that collectively replaces the base-RL velocity-command-tracking rewards and the angular-momentum regularizer by supervising the centroidal-MPC's predicted CoM trajectory under the prediction-confidence weighting schedule of Eq.~\eqref{eq:mpc_com_landmark_annealing};
\texttt{mpc\_ang\_mom} regulates angular momentum against the MPC reference;
\texttt{mpc\_foot} adds a spatial signal for foot placement against the
Raibert-heuristic landing targets; and a small \texttt{mpc\_grf}
coefficient softly aligns the foot ground-reaction forces with the
MPC's optimal solution, bridging the centroidal abstraction and the
full-body dynamics.  The shared regularization block is held identical to the base-RL configuration, so that any locomotion-quality difference between the two reward stacks is
attributable to the MPC-guided block alone.

\subsection{Effective-Inertia-Based Joint PD Gain Selection}
For training and hardware deployment on Themis, we select joint PD gains based on the effective inertia observed at each joint. Around a nominal configuration $q^\star$, the local joint dynamics are approximated as
\begin{equation}
J_{\mathrm{eff},i}\ddot{q}_i + K_{d,i}\dot{q}_i + K_{p,i}q_i \approx 0,
\label{eq:joint_dyn}
\end{equation}
\begin{equation}
J_{\mathrm{eff},i} \approx M_{ii}(q^\star) + A_i.
\end{equation}
Here, $M_{ii}(q^\star)$ is the diagonal rigid-body inertia seen at joint $i$, including downstream link effects, and $A_i$ denotes the joint armature, including motor-reflected inertia and actuator-side effects.
Matching \eqref{eq:joint_dyn} to a second-order system gives
\begin{equation}
K_{p,i}=J_{\mathrm{eff},i}\omega_{n,i}^2,
\qquad
K_{d,i}=2\zeta_iJ_{\mathrm{eff},i}\omega_{n,i},
\end{equation}
where $\omega_{n,i}$ is the desired natural frequency and $\zeta_i$ is the damping ratio.

Unlike reflected-inertia-only heuristics \cite{liao2025beyondmimic} that set gains from $A_i$ alone with a fixed high natural frequency, we include the downstream inertia term $M_{ii}(q^\star)$ and intentionally choose a low $\omega_{n,i}$. This is necessary for Themis because its high-inertia, low gear-ratio actuators do not perform as stiff servo-like motors (\textit{e.g.,} G1), and can excite unmodeled actuator compliance when high-bandwidth gains are used. Therefore, carefully selecting a favorable actuator natural frequency based on actuator behavior is key (1-2 Hz for our case). The resulting lower-bandwidth impedance improves hardware stability and avoids severe chatter at the hip and knee joints. A motor gain comparison is shown in Table \ref{tab:themis_gain_selection}.

\section{Results}
\label{sec:Results}

\begin{figure}[!t]
\vspace{0.15cm}
    \center	
    \includegraphics[clip, trim=0.2cm 2cm 0.1cm 1.6cm, width=1\columnwidth]{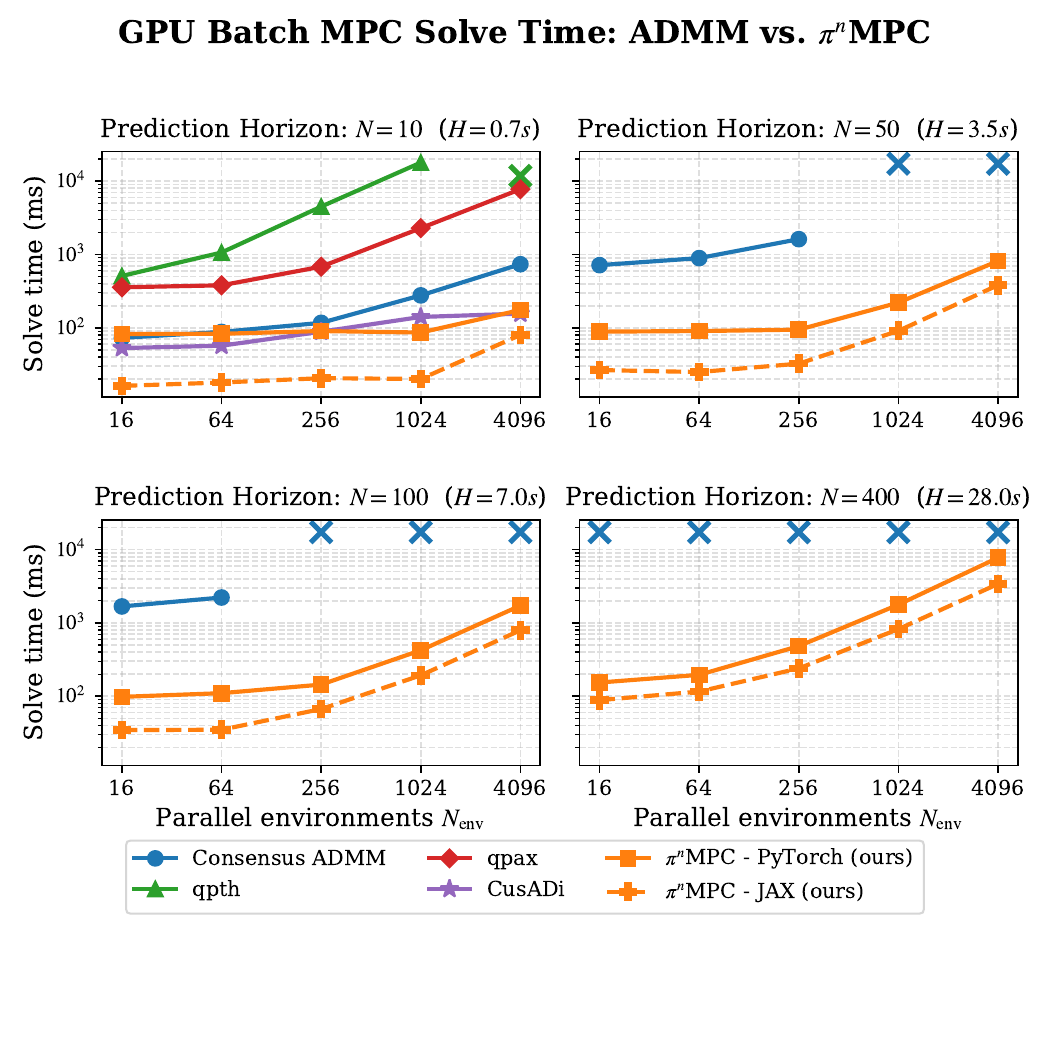}
    \caption{Scalability of Batched GPU MPC Solvers. ($\times$ means CUDA out of memory; Number of max iteration is 200 for consensus ADMM and $\pi^n$MPC, 20 for qpth, 30 for qpax, and 5 PDIPM iterations for CusADi. Note CusADi compilation takes 1.5 hr and requires 150 GB RAM for $N=10$.)}
    \label{fig:scalibility}
    \vspace{-0.3cm}
\end{figure}

\definecolor{MPCReward}{RGB}{255,244,226}
\newcolumntype{G}{>{\centering\arraybackslash}p{8pt}}

\definecolor{MPCReward}{RGB}{255,244,226}
\newcolumntype{G}{>{\centering\arraybackslash}p{4pt}}

\begin{table*}[t]
\centering
\footnotesize
\caption{Time-varying velocity tracking, push-recovery, and constraint-satisfaction comparison between Pure RL and MPC-RL under different MPC horizons and update rates. ($\uparrow$ high is better, $\downarrow$ lower is better.)
}
\label{tab:mpc_rl_tracking_push}
\setlength{\tabcolsep}{2.5pt}
\renewcommand{\arraystretch}{1.15}
\begin{adjustbox}{max width=\textwidth}
\begin{tabular}{l c c c c c G c c c G c c c c c c c c G c c c}
\toprule
\multirow{2}{*}{Method}
& \multicolumn{2}{c}{Training-time MPC}
& \multicolumn{7}{c}{Time-varying velocity tracking RMSE $\downarrow$}
&
& \multicolumn{8}{c}{Max recoverable push force (N) $\uparrow$}
&
& \multicolumn{3}{c}{Constraint satisfaction $\uparrow$} \\
\cmidrule(lr){2-3}
\cmidrule(lr){4-10}
\cmidrule(lr){12-19}
\cmidrule(lr){21-23}
& $N$ & Rate
& $v_x$ & $v_y$ & $\omega_z$
&
& $v_x^{\mathrm{OOD}}$ & $v_y^{\mathrm{OOD}}$ & $\omega_z^{\mathrm{OOD}}$
&
& $0^\circ$ & $45^\circ$ & $90^\circ$ & $135^\circ$
& $180^\circ$ & $225^\circ$ & $270^\circ$ & $315^\circ$
&
& Friction & CWC-$y$ & CWC-$x$ \\
\midrule

Pure RL
& -- & --
& 0.1609 & 0.1566 & 0.0912
&
& 0.2310 & 0.1975 & \textbf{0.1113}
&
& 300 & 350 & 100 & 300 & 350 & 225 & 125 & 275
&
& 100$\%$ & 57.2$\%$ & 92.0$\%$ \\

\arrayrulecolor{gray!45}
\hline
\arrayrulecolor{black}

\multirow{6}{*}{MPC-RL}
& 10 & 5 Hz
& 0.1374 & 0.1342 & 0.1060
&
& 0.2052 & 0.1851 & 0.1138
&
& 325 & 375 & 75 & 325 & 425 & 325 & 75 & 125
&
& 100$\%$ & 69.8$\%$ & 98.2$\%$ \\

\arrayrulecolor{gray!45}
\cline{2-23}
\arrayrulecolor{black}

& \cellcolor{MPCReward}10
& \cellcolor{MPCReward}10 Hz
& \cellcolor{MPCReward}\textbf{0.1280}
& \cellcolor{MPCReward}\textbf{0.1326}
& \cellcolor{MPCReward}0.0928
& \cellcolor{MPCReward}
& \cellcolor{MPCReward}\textbf{0.1949}
& \cellcolor{MPCReward}\textbf{0.1761}
& \cellcolor{MPCReward}0.1171
& \cellcolor{MPCReward}
& \cellcolor{MPCReward}375
& \cellcolor{MPCReward}375
& \cellcolor{MPCReward}125
& \cellcolor{MPCReward}\textbf{400}
& \cellcolor{MPCReward}\textbf{450}
& \cellcolor{MPCReward}\textbf{375}
& \cellcolor{MPCReward}100
& \cellcolor{MPCReward}350
& \cellcolor{MPCReward}
& \cellcolor{MPCReward}100$\%$
& \cellcolor{MPCReward}\textbf{91.8$\%$}
& \cellcolor{MPCReward}94.0$\%$ \\

\arrayrulecolor{gray!45}
\cline{2-23}
\arrayrulecolor{black}

& 10 & 25 Hz
& 0.1340 & 0.1455 & 0.1058
&
& 0.1977 & 0.1868 & 0.1399
&
& \textbf{425} & 375 & 175 & 250 & 350 & 300 & 200 & 375
&
& 100$\%$ & 84.8$\%$ & 94.8$\%$ \\

\arrayrulecolor{gray!45}
\cline{2-23}
\arrayrulecolor{black}

& 10 & 50 Hz
& 0.1464 & 0.1375 & 0.1050
&
& 0.2049 & 0.1811 & 0.1411
&
& 400 & 375 & 150 & 350 & 400 & 325 & 200 & \textbf{400}
&
& 100$\%$ & 82.0$\%$ & 98.2$\%$ \\

\arrayrulecolor{gray!75}
\cline{2-23}
\arrayrulecolor{black}

& 30 & 10 Hz
& 0.1438 & 0.1337 & \textbf{0.0873}
&
& 0.2109 & 0.1834 & 0.1374
&
& 325 & \textbf{400} & 175 & 275 & 400 & 350 & 125 & 300
&
& 100$\%$ & 87.8$\%$ & \textbf{99.0$\%$} \\

\arrayrulecolor{gray!45}
\cline{2-23}
\arrayrulecolor{black}

& 50 & 10 Hz
& 0.1316 & 0.1511 & 0.0923
&
& 0.1985 & 0.1842 & 0.1259
&
& 325 & 350 & \textbf{200} & 350 & 375 & 325 & \textbf{300} & 350
&
& 100$\%$ & 81.0$\%$ & 96.0$\%$ \\

\bottomrule
\multicolumn{23}{@{}l@{}}{\footnotesize The highlighted row is the default MPC-RL configuration, $N=10$ at $10$ Hz. OOD command bounds are 50$\%$ higher than the in-distribution commands.}
\end{tabular}
\end{adjustbox}
\end{table*}

\begin{figure*}[!t]
    \centering
    \begin{subfigure}[t]{0.48\textwidth}
        \centering
        \includegraphics[clip, trim=0.3cm 0.2cm 0.2cm 0.2cm, width=\textwidth]{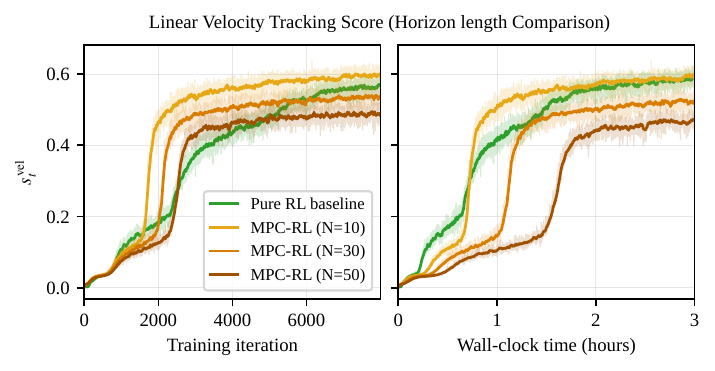}
        \caption{Varying MPC training-time horizon length (all runs are at 10 Hz rate).}
        \label{fig:lin_vel_score_horizon}
    \end{subfigure}
    \begin{subfigure}[t]{0.48\textwidth}
        \centering
        \includegraphics[clip, trim=0.3cm 0.2cm 0.2cm 0.2cm, width=\textwidth]{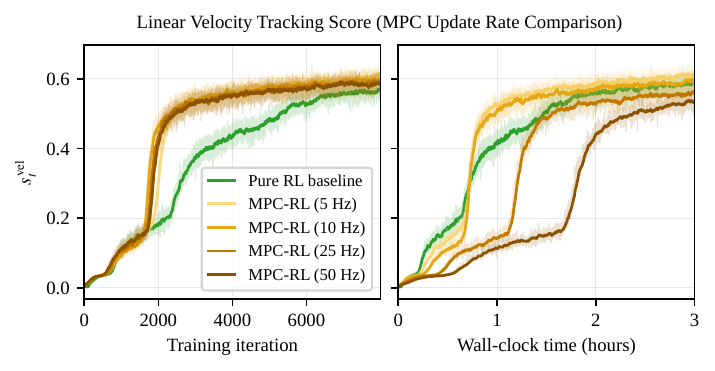}
        \caption{Varying MPC training-time update rate (all runs have fixed $N=10$).}
        \label{fig:lin_vel_score_rate}
    \end{subfigure}
    \caption{Comparison of locomotion velocity-tracking score metric vs. training steps and wall-clock time under different MPC-RL settings.}
    \label{fig:lin_vel_combined}
    \vspace{-0.2cm}
\end{figure*}

This section evaluates the proposed MPC-guided Reinforcement Learning (MPC-RL for short) framework from three perspectives. We first characterize the computational scalability of $\pi^n$MPC inside massively parallel training. We then study the effect of MPC guidance on humanoid locomotion through empirical comparisons. Finally, we extend to a loco-manipulation task involving payload pushing.

\subsection{Experimental Setup}
\label{sec:exp_setup}

The desktop setup for training and evaluation consists of an Intel Ultra 9 285k CPU,  an Nvidia RTX 5090 GPU (32GB VRAM), and 64 GB of RAM.
The trained policy receives the same observation space across all ablations; only the reward guidance or MPC configuration is varied.
We evaluate the locomotion and loco-manipulation policies on the Westwood Robotics Themis V2 humanoid robot with custom Python-based deployment infrastructure. 

\subsection{Acceleration and Scalability of $\pi^n$MPC}
\label{sec:solver_benchmark}

We first evaluate the efficiency and scalability of $\pi^n$MPC on the same CD-MPC problem used during training with SOTA methods, including \texttt{qpax}, a JAX-based batch QP solver \cite{qpax}, \texttt{qpth}, a PyTorch-based batch QP solver \cite{qpth}, CusADi, a symbolic GPU solver \cite{jeon2024cusadi}, and consensus ADMM, a custom batched implementation of \cite{neal2011distributed}, with QP-splitting structure following \texttt{OSQP} \cite{stellato2018osqp}. As shown in Fig.~\ref{fig:scalibility}, in the $N=10$ case, $\pi^n$MPC scales more favorably on GPU than consensus ADMM, \texttt{qpth}, and \texttt{qpax} as the number of parallel environments increases. CusADi shares comparable solve speed, but its symbolic compilation takes around 1.5 hours for our MPC problem.
Our advantage is especially clear for longer horizons, where \texttt{qpth} and \texttt{qpax} are unable to provide competitive solve time for comparison here; CusADi requires substantially longer compilation time ($>$5 hrs) and high RAM usage, and thus are omitted from the rest of the plots. Consensus ADMM either becomes substantially slower or runs out of VRAM, while $\pi^n$MPC remains feasible by parallelizing over both environments and horizon steps even to $N=400$. 

Note that $\pi^n$MPC's JAX implementation is generally more computationally efficient than the PyTorch implementation; however, in RL training, $\pi^n$MPC-JAX's memory overhead is sizably larger than the PyTorch variant due to the memory splitting and pre-allocation.

\subsection{Humanoid Locomotion with MPC-RL}
\label{sec:locomotion_results}

Having established that long-horizon CD-MPC can be computed at training scale, we next study how this predictive guidance can improve humanoid locomotion learning. Here, we set MPC $\Delta t=0.7$s, such that $5\Delta t$ makes up one stride.

\subsubsection{Choosing the right MPC setup} 
First, we study how the training-time MPC update rate and prediction horizon affect the final locomotion policy. Table~\ref{tab:mpc_rl_tracking_push} shows that MPC-RL consistently improves translational velocity tracking over Pure RL across velocity command tracking tests, including out-of-distribution commands. MPC-RL variants also outperform the RL baseline in all tested directions under push-recovery experiments.
We additionally evaluate the contact constraint satisfaction during deployment, training-time MPC's friction cone and CWC constraints, and their shaping to GRF guidance leads to improved constraint satisfaction over base RL's \texttt{foot\_slip} penalty alone.

We compare the effects of training-time MPC horizon number with a fixed 10 Hz update rate. Although longer horizons provide more lookahead, Table~\ref{tab:mpc_rl_tracking_push} and Fig.~\ref{fig:lin_vel_score_horizon} show that performance does not improve monotonically with horizon length; the $N=10$ setting (one complete gait cycle lookahead) gives the strongest overall velocity-tracking and push-recovery behavior, while maintaining a very competitive training efficiency compared to baseline RL.
We further compare training-time MPC update rates at a fixed horizon $N=10$. The $10$ Hz setting achieves the best overall tracking performance with good training efficiency, while higher-rate guidance does not further improve the learned policy. This trend is also reflected in Fig.~\ref{fig:lin_vel_score_rate}, where increasing the MPC rate exhibits diminishing returns on policy performance. 

\textit{Therefore, we use N=10, 10 Hz training-time MPC setup for the following experiments in this paper. The locomotion policy is trained to 15k iterations under 6 hours.}

\subsubsection{Comparison of Model-based Guidance Reward Design}

Finally, we compare the model-free baseline and model-based reward structures seen in the current literature. Comparing (i) baseline pure RL reward, (ii) constructing MPC landmarks as CLF-based reward \cite{li2026clf}, (iii) interpolated oracle guidance reward \cite{krishna2024ogmp}, and (iv) the proposed MPC-guided reward. We additionally compare the choice of $\alpha$ in the MPC landmark schedule, including ``next-step" $\{1,~0,~0,~0,~0\}$, ``averaged" $\{0.2,~0.2,~0.2,~0.2,~0.2\}$, and the proposed ``tapered" $\{0.5,~0.25,~0.15,~0.07,~0.03\}$.

As Table \ref{tab:locomotion_method_ablation} suggests, the proposed MPC-guided reward structure with tapered landmark weighting has clear advantages in locomotion velocity and push recovery performance against the other model-guided reward design approaches. We compare the progress of the linear velocity tracking score during training, and the MPC-RL achieves faster progress (higher $\%$ in unit time) even compared to the pure RL approach, despite parallel MPCs adding slightly more training iteration time. The proposed horizon-wise tapered tracking schedule works more favorably than the immediate next-step or average-weighted tracking schedule.


\begin{table}[!t]
\vspace{0.15cm}
\centering
\footnotesize
\caption{Performance comparison of model/trajectory-guided RL methods for humanoid locomotion.}
\label{tab:locomotion_method_ablation}
\setlength{\tabcolsep}{2pt}
\renewcommand{\arraystretch}{1.15}
\begin{adjustbox}{max width=\columnwidth}
\begin{tabular}{l c c c c c c c c c}
\toprule
\multirow{2}{*}{\makecell{Reward\\Method}}
& \multicolumn{2}{c}{Tracking RMSE $\downarrow$}
& \multicolumn{4}{c}{Max recovery force (N) $\uparrow$}
& \multicolumn{3}{c}{$s_t^\textrm{vel}/s_\textrm{max}^\textrm{vel}$ $^\dagger$ ($\%$) $\uparrow$} \\
\cmidrule(lr){2-3}\cmidrule(lr){4-7}\cmidrule(lr){8-10}
&
\makecell{$v_x$}
& \makecell{$v_y$}
& \makecell{$0^\circ$}
& \makecell{$90^\circ$}
& \makecell{$180^\circ$}
& \makecell{$270^\circ$}
& \makecell{30min}
& \makecell{45min}
& \makecell{1hr} \\
\midrule

Baseline RL
& 0.161 & 0.157 & 300 & 100 & 350 & 125 & \textbf{24.3} & 52.6 & 63.7 \\

CLF-RL \cite{li2026clf}
& 0.151 & 0.139 & 375 & 125 & 350 & 150 & 11.8 & 18.9 & 54.9 \\

OGMP \cite{krishna2024ogmp}
& 0.148 & 0.161 & 375 & \textbf{200} & 375 & 150 & 22.9 & 26.7 & 33.0 \\

\arrayrulecolor{gray!65}
\midrule
\arrayrulecolor{black}

MPC-RL
& \multicolumn{9}{c}{} \\

\arrayrulecolor{gray!35}
\cline{2-10}
\arrayrulecolor{black}

\quad Next-step
& 0.190 & 0.170 & 325 & 125 & 350 & 100 & 18.7 & 25.0 & 68.7 \\

\quad Averaged
& 0.134 & 0.136 & \textbf{425} & 150 & 350 & 150 & 21.9 & 28.2 & 71.5 \\

\quad \textbf{Tapered}
& \textbf{0.128} & \textbf{0.132} & 400 & 150 & \textbf{450} & \textbf{250} & 16.7 & \textbf{60.0} & \textbf{77.7} \\

\bottomrule
\multicolumn{10}{@{}l@{}}{\footnotesize
$\uparrow$ high is better, $\downarrow$ lower is better.} \\
\multicolumn{10}{@{}l@{}}{\footnotesize
$^\dagger$ $\%$ progress of a unified linear velocity tracking score metric during training.}
\end{tabular}
\end{adjustbox}
\vspace{-0.5cm}
\end{table}

We validate the proposed MPC-RL policy on hardware -- as Figure \ref{fig:hardware}(a-e) presents, we demonstrate MPC-RL's hardware transferability and robustness via unknown disturbances, including external weights and pushes.


\subsection{Humnoid Loco-Manipulation with MPC-RL}
\label{sec:loco_manipulation}
We further evaluate the proposed MPC-RL framework on a humanoid box-pushing task, using a loco-manipulation MPC at training time to generate additional hand-pushing force references and box state references.
Compared with the pure RL baseline, the structured MPC rewards provide more informative exploration signals by guiding the policy toward dynamically consistent CoM, momentum, and contact-force behaviors during pushing. 

\begin{figure}[!t]
    \centering
    \vspace{0.15cm}
    \begin{subfigure}[t]{0.45\textwidth}
        \centering
        \includegraphics[clip, trim=0.0cm 0.0cm 0.0cm 0.0cm, width=\textwidth]{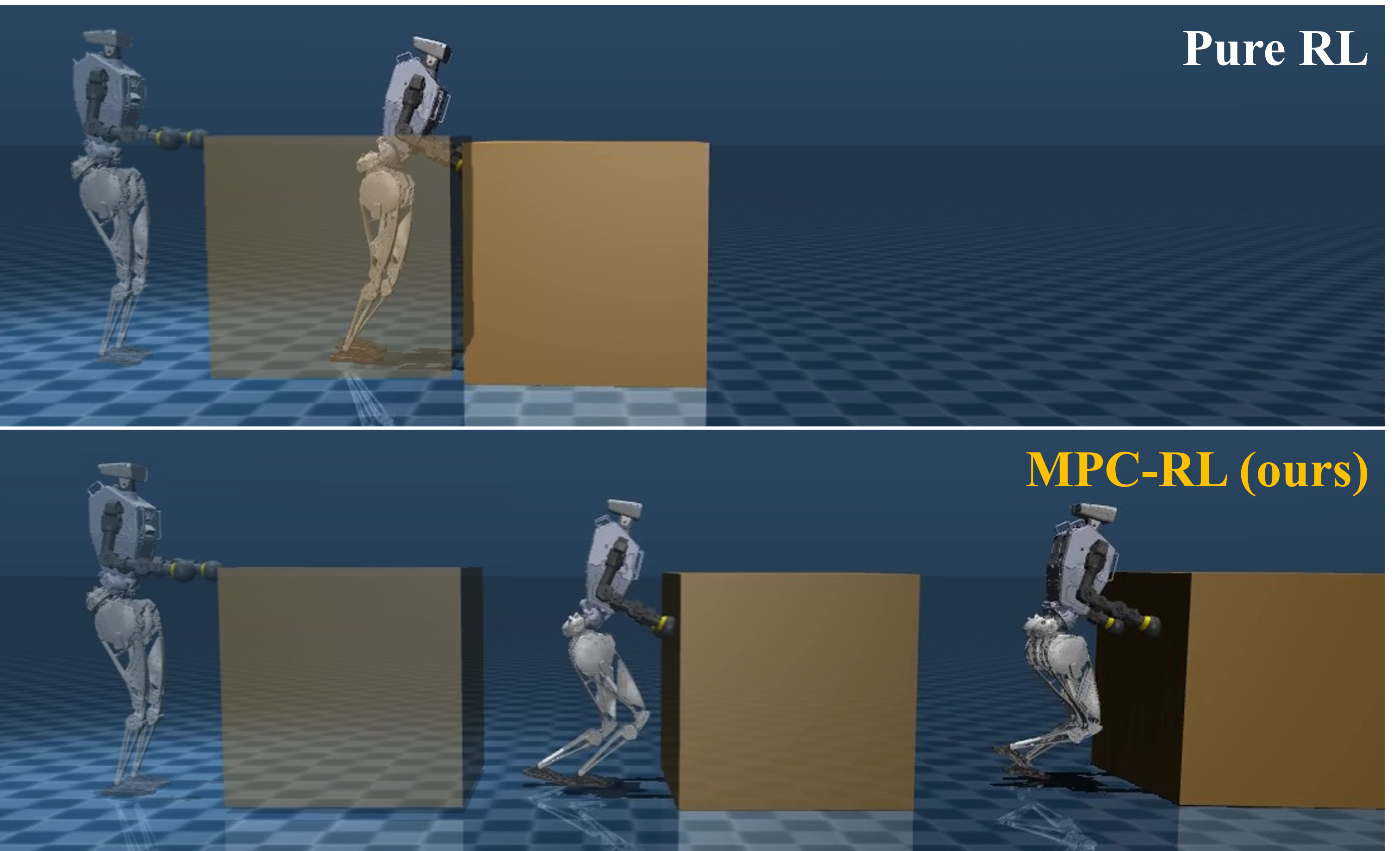}
        \caption{Simulation snapshots.}
        \label{fig:sim_pushbox}
    \end{subfigure}
    \begin{subfigure}[t]{0.48\textwidth}
        \includegraphics[clip, trim=0.2cm 0.2cm 0.1cm 0.1cm, width=1\columnwidth]{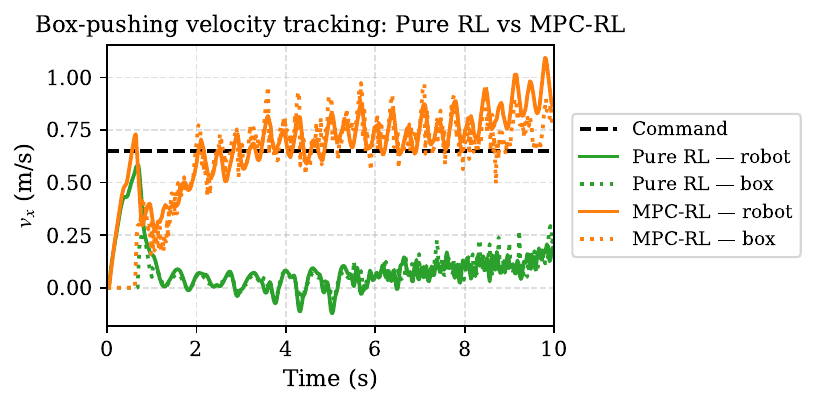}
        \caption{Velocity tracking plots.}
    \label{fig:pushing_plot}
    \end{subfigure}
    \caption{Velocity tracking comparison in box-pushing simulation. The robot is commanded to push a 20kg, 1m$\times$1m$\times$1m box with a contact friction coefficient of 0.2.}
    \label{fig:pushing}
    \vspace{-0.2cm}
\end{figure}

We compare the standard pure RL training vs. MPC-RL in this setting. Both trainings involve boxes with random masses in the training environment. Pure RL adds box and body velocity tracking rewards, body-box velocity matching rewards, and contact sticking rewards to incentivize pushing, whereas MPC bundles centroidal state and reaction force predictions into a structured MPC reward.
As a result, as shown in Figure \ref{fig:pushing}, the MPC-guided policy achieves better velocity tracking and stronger pushing capability over the baseline, which stalls from the beginning. 
The loco-manipulation policy uses the same observation space as the locomotion MPC-RL policy and does not require runtime feedback of object states or parameters, unlike pure MPC approaches.

We validate this loco-manipulation policy on hardware, as shown in Figure~\ref{fig:hardware}(f), the robot is capable of pushing up to 290 kg (639 lbs) payload, which is equivalent to 829$\%$ body mass. The measured pushing force required for this cart is about 180 N. 
These results suggest that training-time MPC guidance is not limited to locomotion tracking but can also provide useful physical structure for discovering forceful loco-manipulation, where pure RL suffers from excessive exploration.

\section{Conclusion and Future Perspectives}
\label{sec:Conclusion}

This work presented an efficient MPC-guided reinforcement learning framework for humanoid locomotion and loco-manipulation. By combining prediction-confidence weighting with $\pi^n$MPC, the proposed method provides structured centroidal-dynamics guidance during training while avoiding test-time MPC. The results show improved velocity tracking, push-recovery robustness, and reward progression over pure RL and model-guided baselines, with practical training overhead. We further demonstrated that the same framework can guide forceful loco-manipulation, enabling hardware pushing of heavy payloads over 600 lbs on a cart.

We intend to use this work as a first step to showcase the validity and significance of structured MPC guidance in RL. There are still limitations in the reliance on reduced-order centroidal dynamics with prescribed contact schedules. Additionally, the inherent limitation of numerical solvers still presents scaling limitations regardless. Future work will extend the framework to richer contact modes with nonlinear MPC to account for the robot model and contact interaction with more fidelity. We will investigate using neural-network-augmented batch MPC solvers to enable very efficient training-time nonlinear MPC guidance in humanoid reinforcement learning. 

\balance
\bibliographystyle{ieeetr}
\bibliography{reference.bib}

\appendix


\begin{table}[H]
\vspace{0.0cm}
\centering
\caption{Comparison of inertia-based gain selection for THEMIS v2. Here $k_g$ is the gear ratio, $I_{\mathrm{rotor}}=A_i/k_g^2$ is the rotor inertia, and $\omega_n=2\pi f_n$.}
\label{tab:themis_gain_selection}
\scriptsize
\setlength{\tabcolsep}{2.0pt}
\renewcommand{\arraystretch}{1.15}
\resizebox{\columnwidth}{!}{
\begin{tabular}{llccccccc}
\toprule
\makecell[l]{Joint}
&
\makecell[l]{Method}
&
$k_g$
&
\makecell[c]{$I_{\mathrm{rotor}}$\\$(\mathrm{kg\,m^2})$}
&
\makecell[c]{$A_i$\\$(\mathrm{kg\,m^2})$}
&
\makecell[c]{$J_{\mathrm{eff}}$\\$(\mathrm{kg\,m^2})$}
&
\makecell[c]{$f_n$\\$(\mathrm{Hz})$}
&
$\zeta$
&
\makecell[c]{$K_p/K_d$}
\\
\midrule
\multirow{2}{*}{Ab/Ad}
&
\cite{liao2025beyondmimic}
&
\multirow{2}{*}{$9.00$}
&
\multirow{2}{*}{$9.15{\times}10^{-4}$}
&
\multirow{2}{*}{$0.074$}
&
$0.074$
&
$10.0$
&
$2.0$
&
$292.5/18.6$
\\
&
Ours
&
&
&
&
$0.46$
&
$1.4$
&
$0.6$
&
$35.6/4.8$
\\
\arrayrulecolor{gray!45}
\cline{1-9}
\arrayrulecolor{black}
\multirow{2}{*}{\makecell[l]{Hip}}
&
\cite{liao2025beyondmimic}
&
\multirow{2}{*}{$7.78$}
&
\multirow{2}{*}{$1.90{\times}10^{-3}$}
&
\multirow{2}{*}{$0.1153$}
&
$0.115$
&
$10.0$
&
$2.0$
&
$455.2/29.0$
\\
&
Ours
&
&
&
&
$0.45$
&
$1.4$
&
$0.9$
&
$34.8/7.1$
\\
\arrayrulecolor{gray!45}
\cline{1-9}
\arrayrulecolor{black}
\multirow{2}{*}{\makecell[l]{Knee}}
&
\cite{liao2025beyondmimic}
&
\multirow{2}{*}{$7.78$}
&
\multirow{2}{*}{$1.90{\times}10^{-3}$}
&
\multirow{2}{*}{$0.1153$}
&
$0.115$
&
$10.0$
&
$2.0$
&
$455.2/29.0$
\\
&
Ours
&
&
&
&
$0.25$
&
$1.9$
&
$1.2$
&
$35.6/7.2$
\\
\bottomrule
\end{tabular}
}
\vspace{-0.8em}
\end{table}

\end{document}